\documentclass[11pt,a4paper]{article}
\usepackage[hyperref]{acl2019}
\usepackage{times}
\usepackage{latexsym}
\usepackage[utf8]{inputenc}
\usepackage{amsmath}

\usepackage{url}
\usepackage{enumitem}

\aclfinalcopy 

\newcommand\footurl[1]{\footnote{\url{#1}}}
\newcommand\Tref[1]{Table~\ref{tab:#1}}

\title{Unsupervised Lemmatization as Embeddings-Based Word Clustering}

\author{
Rudolf Rosa \qquad Zdeněk Žabokrtský \\
Charles University, Faculty of Mathematics and Physics, \\
Institute of Formal and Applied Linguistics \\
Malostranské náměstí 25, 118 00 Prague 1, Czech Republic \\
{\tt \{rosa,zabokrtsky\}@ufal.mff.cuni.cz}
}

\date{}

\begin{document}
\maketitle
\begin{abstract}
We focus on the task of unsupervised lemmatization, i.e.\ grouping together inflected forms of one word under one label (a lemma) without the use of annotated training data.
We propose to perform agglomerative clustering of word forms with a novel distance measure.
Our distance measure is based on the observation that inflections of the same word tend to be similar both string-wise and in meaning.
We therefore combine word embedding cosine similarity, serving as a proxy to the meaning similarity,
with Jaro-Winkler edit distance.
Our experiments on 23 languages show our approach to be promising, surpassing the baseline on 23 of the 28 evaluation datasets.
\end{abstract}



\section{Introduction}


The task of lemmatization is to assemble inflections of the same word into one group, represented by a designated form of the word called lemma.
It is a classical NLP task, potentially useful e.g.\ for information retrieval or machine translation.

The standard approach is to use supervised machine learning, exploiting a dataset of word and lemma pairs to train a lemmatizer \cite{straka}. However, such datasets are available roughly for only 1\% of world's languages.
An alternative is to use stemming \cite{lovins1968development,porter2001snowball},
which is typically rule-based, i.e.\ does not need annotated training data, but on the other hand is usually language-specific, and also tends to cluster the word forms too coarsely (many different lemmas may share the same stem).

As we want to cover many languages while keeping inflections of different lemmas in separate groups, we instead
propose to perfom
unsupervised lemmatization
as word form clustering.
%

The first step is to employ
a suitable word form distance measure.
We propose a measure combining (a) string similarity, implemented using edit distance, and (b) meaning similarity, for which we use word embedding similarity as a proxy (Sec.~\ref{sec:sim}).

We then precompute distances of probable inflections of the same words, and cluster them with agglomerative clustering (Sec.~\ref{sec:clust}).
We
leave
the last step of selecting a representant from each of the clusters as its lemma
for future work.

We evaluate our setup on 28 datasets for 23 languages, finding that it outperforms the baseline on 23 of the datasets, but also identifying many of its limitations that yet need to be addressed (Sec.~\ref{sec:eval}).

We make our source codes available together with this paper.





\section{Word form distance measure}
\label{sec:sim}


We propose a word form distance measure
which combines string similarity with word embedding similarity,
designed to assess word forms belonging to the same lemma as more similar than word forms belonging to different lemmas.

\subsection{String similarity}
\label{sec:jw}


For string similarity,
we
use the Jaro-Winkler (JW) edit distance \citep{winkler:1990}
from the \texttt{pyjarowinkler} Python package (implemented as a similarity).%
\footurl{https://pypi.org/project/pyjarowinkler/}
Unlike
\citet{levenshtein1966binary} edit distance,
JW gives
more importance to
the beginnings of the strings than to their ends.
We find this to be advantageous,
as most of the inflection usually happens at the end of the word.%
\footnote{Based on the feature 26A of the WALS database by \citet{wals},
most world's languages, and in particular practically all of the languages
in our data set, have a strongly or weakly suffixing inflectional morphology,
whereas prefixing morphology is rare.}

To compute the edit distance of a pair of strings, we average their JW with JW of their \textit{simplified variants};
the simplification consists of lowercasing,
transliteration to ASCII using the Unidecode library,\footurl{https://pypi.org/project/Unidecode/}
and deletion of non-initial vowels (a e i o u y).
This makes the measure
somewhat softer, paying less attention to differences that tend to have lower importance in our setting.



\subsection{Word embedding similarity}

As shown by \citet{word2vec},
cosine similarity of word embeddings tends to capture
morphological, syntactic, and semantic similarities of words.
This motivates our use of
word embedding similarity as a proxy to word meaning similarity.

We use the FastText word embeddings \cite{fasttext}, which have the additional benefit of employing subword embeddings, thus also implicitly capturing string similarity to some extent.\footnote{In our exploratory experiments on a Czech language dataset, we observed the accuracies to rise by approximately 20 percentage points when we substituted word2vec embeddings with FastText embeddings.}

\subsection{Combined distance measure}
Our distance measure is based on a multiplication of the two similarities shifted to the $[0,1]$ interval:
\begin{equation}
    dist(a,b) = 1 - JW(a,b) \cdot \frac{cos(a,b)+1}{2} \\
\end{equation}

\section{Clustering}
\label{sec:clust}

We apply agglomerative clustering
from Scikit-learn \cite{scikit-learn}, with average linkage.\footnote{%
Average linkage is recommended by the manual; we also tried
single and complete linkage, but the results were worse.}
The algorithm starts by assigning each word form to a separate cluster.
In each step, it then merges the pair of clusters with the lowest average
distance of their elements.
The standard stopping criterion is to preset the final number of clusters to
form. As we
have not thought of a way to estimate the number of lemmas,
we instead stop the
algorithm once the cluster distance raises above a threshold $t$; we use $t=0.4$.


The algorithm only assigns a cluster to word forms that are part of the
training vocabulary.
If we encounter an out-of-vocabulary word form at test time,
we perform a single clustering step with it: we assign it to the closest
cluster if it is closer than $t$, otherwise, we put it into a new cluster.




\subsection{Stem-based hyperclusters}
\label{sec:precomputation}

Theoretically, we would like to compute the distances of all pairs of word forms.
In practice,
with a vocabulary of $10^5$ word forms, 
computing the $10^{10}$ distances
would use prohibitive amounts of time and memory.
%
%
Therefore, we use a stemming approach
to pre-partition the space into hyperclusters,
and
run the clustering algorithm on each such hypercluster separately;
word forms with different stems thus cannot be clustered into the same cluster.
%
In this work, we
define the stem of a word form as the first $K$
characters of its \textit{simplified variant} (see sec.~\ref{sec:jw});
we use $K=3$.\footnote{%
This reduces the time and memory complexity roughly 1,000 times; each experiment then uses about 1 GB and 1 h.}



Such crude stemming obviously separates some forms of the same lemma into separate
hyperclusters (short words, irregular inflections, suppletives\dots), making it impossible for our method to reach the correct
clustering.
We intend to address this more properly in future work, especially by
utilizing unsupervised morphological splitting
to get better stems.
However, some of the weak points of our approach, such as suppletives,
probably cannot be easily resolved.



\section{Experiments}
\label{sec:eval}

\subsection{Data}

\begin{table}
    \centering
    \begin{tabular}{r|r}
    Vocabulary size & OOV rate \\
    \hline
        1,000 & 50.3\% \\
       10,000 & 27.6\% \\
      100,000 &  8.8\% \\
    1,000,000 &  1.5\% \\
    \end{tabular}
    \caption{Proportion of test-data word forms that are not part of the vocabulary, for cs\_pdt.}
    \label{tab:oovs}
\end{table}

For the experiments reported in this paper, we use the pretrained word embedding dictionaries available from the FastText website.\footurl{https://fasttext.cc/}%
\footnote{A drawback of using pre-trained word embeddings is that they
typically use different tokenization than the evaluation data, necessarily leading to occasional problems.}
The word embeddings had been trained on Wikipedia\footurl{https://www.wikipedia.org/} and Common Crawl\footurl{http://commoncrawl.org/} texts with the FastText tool,
``using CBOW with position-weights, in dimension 300, with character n-grams of length 5, a window of size 5 and 10 negatives''.
We limit our vocabulary to $N$ most frequent words, i.e.\ the first $N$ words stored in the embedding dictionary; we use $N = 100,000$.
We found that with a smaller dictionary, the method is more efficient computationally, but the results are worse due to very high rates of out-of-vocabulary items (see \Tref{oovs}).

We evaluate on
treebanks from the Universal Dependencies 2.3 \cite{ud23}.
We use the \textit{dev} part for evaluation,
assigning each of the words in this part of the treebank to a cluster and then evaluating the clusters against the gold-standard lemmas
(repeated word forms are used repeatedly in the evaluation, i.e.\ the evaluation is token-based instead of type-based).
We used only treebanks that satisfy the following criteria:
\begin{itemize}[nosep]
    \item Contain at least 100,000 tokens.
    \item Lemmas are annotated (semi-)manually.%
    \footnote{I.e.\ the ``Lemmas'' feature in the treebank Readme file is ``manual native'', ``converted from manual'', ``converted with corrections'', or ``automatic with corrections''.}
    \item There are pretrained FastText embeddings available for the language.
\end{itemize}
This results in a set of 28 treebanks for 23 languages (a subset of the total 129 treebanks for 76 languages), listed in \Tref{results}.
For more information on the treebanks, please consult the UD webpage.\footurl{https://universaldependencies.org/}

We used the cs\_pdt treebank to tune the method and set its hyperparameters.



\subsection{Evaluation}

We evaluate the clustering using the standard Scikit \textit{v-measure},
which penalizes both clusters containing forms of multiple lemmas as well as forms of a single lemma scattered in multiple clusters.
The results are listed in Table~\ref{tab:results}.


As a baseline, we choose the better-performing of these two approaches for each dataset:
either taking the form as the lemma, or taking the first 5 characters of the form as the lemma.\footnote{%
For languages with little inflection, using the form is usually better;
for highly inflectional languages,
the 5 character prefix usually performs better.}

The upper bound is an oracle, always selecting the gold standard lemma if it is located in the same hypercluster.
Due to the stem-based hyperclustering (sec.~\ref{sec:precomputation}), the oracle does not reach 100\%;
this constitutes one of the strongest limitations of our approach.\footnote{%
We have tried to selectively remerge at least some of the hyperclusters by using a coarser but more efficient merging strategy; however, we have not been successful with this approach so far.}


We also express the performance of our method as error reduction on the scale from baseline (0\%) to upper bound (100\%).


\begin{table}
    \centering
    \small
    \begin{tabular}{l|lrrr|r}
        \hline
Treebank
& \multicolumn{2}{l}{Baseline}
& \multicolumn{1}{l}{Our}
& \multicolumn{1}{l|}{Upp.}
& \multicolumn{1}{l}{Err.red.} \\
        \hline
        ar\_padt & form & 4.19 & 3.90 & 2.93 & 23.1 \\
ca\_ancora & form & 4.65 & 4.35 & 3.32 & 22.3 \\
cs\_cac & form5 & 3.56 & 2.25 & 1.14 & 54.0 \\
cs\_fictree & form5 & 4.82 & 4.08 & 2.68 & 34.6 \\
\hline
cs\_pdt & form5 & 4.93 & 3.41 & 1.65 & 46.6 \\
da\_ddt & form & 2.32 & 2.16 & 1.55 & 21.2 \\
en\_ewt & form & 2.29 & 2.22 & 1.78 & 13.8 \\
es\_ancora & form & 3.99 & 3.38 & 2.25 & 34.7 \\
\hline
et\_edt & form5 & 4.78 & 4.31 & 2.54 & 20.9 \\
fa\_seraji & form & 8.99 & 8.76 & 7.44 & 14.8 \\
fr\_gsd & form & 4.12 & 3.81 & 2.70 & 22.0 \\
hi\_hdtb & form & 4.18 & 3.58 & 2.83 & 44.3 \\
\hline
hr\_set & form5 & 4.04 & 2.87 & 1.71 & 50.2 \\
it\_isdt & form & 4.27 & 3.71 & 2.78 & 37.8 \\
it\_postwita & form & 3.60 & 4.07 & 2.37 & -38.0 \\
ja\_gsd & form & 1.64 & 1.93 & 1.41 & -123.1 \\
\hline
ko\_kaist & form & 0.14 & 2.41 & 0.11 & -6392.8 \\
la\_ittb & form5 & 6.53 & 6.97 & 3.85 & -16.4 \\
la\_proiel & form5 & 6.92 & 7.42 & 4.20 & -18.4 \\
lv\_lvtb & form5 & 3.90 & 3.39 & 2.10 & 28.0 \\
\hline
no\_bokmaal & form & 2.79 & 2.22 & 1.48 & 43.6 \\
no\_nynorsk & form & 2.73 & 2.52 & 1.48 & 16.7 \\
pl\_lfg & form5 & 3.68 & 3.06 & 1.84 & 33.6 \\
pt\_bosque & form & 3.57 & 3.17 & 2.55 & 39.0 \\
\hline
ro\_nonstd & form5 & 8.13 & 7.95 & 5.64 & 7.2 \\
sk\_snk & form5 & 2.87 & 2.01 & 0.63 & 38.2 \\
uk\_iu & form & 2.66 & 1.94 & 0.88 & 40.7 \\
ur\_udtb & form & 3.95 & 3.79 & 2.65 & 12.3 \\
\hline
Average &  & 4.08 & 3.77 & 2.45 & -210.3 \\
Median &  & 3.97 & 3.40 & 2.31 & 22.7 \\

        \hline
    \end{tabular}
    \caption{Results of form clustering, measured in \% of $1 - vmeasure$ (expressing the error, i.e.\ lower is better). Baseline (either full form or prefix of form of length 5), our system, and oracle upper bound. Last column is error reduction on the scale from baseline to upper bound, in \%.}
    \label{tab:results}
\end{table}

\subsection{Discussion}

For 23 of the 28 datasets, our method achieves a positive error reduction; the median error reduction is 23\%.
Because of the extreme result for Korean, the average does not make much sense here.

The results are worst for Korean and Japanese, which are analytical languages with practically no inflection, making the ``form'' baseline very close to the upper bound -- the clusters should mostly have the size of 1. As our hyperparameters are not tuned for this, and the whole idea of lemmatizing these languages is questionable, our results are very low here.
We also observe deteriorations for a treebank of Italian Tweets and for treebanks of historical Latin, which are all known to be very hard datasets.

On all other datasets, we observe an improvement over the baseline, with an error reduction typically between 10\% and 35\%.
The performance is especially good for Slavic languages (cs, hr, pl, sk, uk), where the error reduction is often around 50\%. This is most probably due to the hyperparameters being tuned on the cs\_pdt treebank.



The threshold $t$ controls the balance between too conservative and too eager cluster merging.
Being too conservative typically leaves some string-wise distant inflections, such as different verb tenses, in separate clusters.
Being too eager tends to also merge word forms related by derivation rather than inflection, e.g.\ adjectives and their corresponding adverbs.
We have tried to avoid such merges by using part-of-speech (POS) tags to further separate the word forms, which seemed promising with supervised POS.
However, we do not want to rely on supervised POS in an unsupervised method, and we observed poor results once we moved to the (rather noisy) unsupervised POS of \citet{deltacorpus}.

We are convinced that the proposed approach is still too weak, lacking the means to reliably separate true inflections from other similar word forms. For this, we believe, it will be necessary to actively look for regularities in the clusters to try to recognize inflectional paradigms in them, and then
refine the clusters by encouraging them to match the paradigms.

As the assignment of word forms to clusters is context-independent, the approach also cannot deal with homonymy.
We believe this could be solved by switching from context-independent word embeddings to contextual word embeddings, such as
BERT \cite{bert}.

\begin{table}
    \centering
    \small
    \begin{tabular}{lrr}
        \hline
        Distance   & \multicolumn{1}{c}{Average} & \multicolumn{1}{c}{Median} \\
        \hline
        $JW$     & 8.17 & 7.92 \\
$cos$    & 4.39 & 3.87 \\
$JW\cdot cos$ & 3.77 & 3.40 \\ 

        \hline
    \end{tabular}
    \caption{Comparison of the word form similarities, in \% of $1 - vmeasure$ of the clustering. Average and median over the 28 datasets.}
    \label{tab:avg}
\end{table}

In Table~\ref{tab:avg}, we compare the combined distance measure with each of the two components used alone.
The results show that combining the edit distance with the embedding similarity is stronger than using any of the measures alone.
In fact, only for two datasets, $cos$ was slightly better than $JW\cdot cos$.
The embedding similarity alone performs much better than the edit distance alone.
We believe that this is at least partially due to FastText embeddings' incorporation of subword information, which means that their similarity also captures string similarity to some extent.

\section{Conclusion}

In this work, we have suggested an approach to unsupervised lemmatization based on agglomerative clustering,
using a word form distance measure combining string similarity (edit distance) and meaning similarity (word embeddings).
The evaluation showed the approach to be promising, surpassing the baseline on most of the evaluation datasets.
At the same time, it has many obvious weak points that need to be addressed in future.



\bibliography{acl2019}
\bibliographystyle{acl_natbib}

\end{document}